%% file: main.tex
\documentclass[11pt, logo, onecolumn, copyright, colorlinks=true, allcolors=blue]{nvidiatechreport}
\usepackage{graphicx}
\usepackage{subcaption,ragged2e}
\usepackage[round]{natbib}

\usepackage[normalem]{ulem}
\usepackage[utf8]{inputenc}
\usepackage[T1]{fontenc}
\usepackage{url}
\usepackage{tikz}
\usepackage{enumitem}
\usetikzlibrary{positioning}
\usetikzlibrary{calc}

\usepackage{array}
\usepackage{booktabs}
\usepackage{wrapfig}
\usepackage{caption}
\usepackage{listings}
\usepackage{adjustbox}
\usepackage{footnote}
\usepackage{multirow}
\usepackage{amsmath}
\usepackage{amsfonts}
\usepackage{amssymb}
\usepackage{breakcites}
\usepackage{svg}
\usepackage[most]{tcolorbox}
\usepackage{tabularx}
\usepackage{xcolor}
\usepackage{colortbl}
\usepackage{algorithm}
\usepackage{siunitx}
\usepackage{algorithmic}
\usepackage{xspace}
\newcommand{\newparagraph}[1]{\noindent\textbf{#1\hspace{0.5em}}\ignorespaces}

\newcommand{\remove}[1]{}

\newcommand{\ourmethod}{TwoTower\xspace}

\newcommand{\ourmodel}{Nemotron-Labs-TwoTower\xspace}
\newcommand{\backbone}{Nemotron-3-Nano-30B-A3B\xspace}
\newcommand{\backboneshort}{Nemotron-3-Nano\xspace}
\newcommand{\mask}{\texttt{[MASK]}\xspace}
\title{Nemotron-Labs-TwoTower: Diffusion Language Modeling with Pretrained Autoregressive Context}
\author{Fitsum Reda$^{*\dagger}$, John Kamalu$^{*}$, Roger Waleffe, Mostofa Patwary, Mohammad Shoeybi, Bryan~Catanzaro}
\date{}

\begin{document}

\begin{abstract}
\large \textbf{Abstract}
\input{sections/abstract.tex}
\end{abstract}

\maketitle
\begingroup
\renewcommand{\thefootnote}{*}
\footnotetext{Core contributor.}
\renewcommand{\thefootnote}{\ensuremath{\dagger}}
\footnotetext{Project lead. Contact: \texttt{freda@nvidia.com}}
\endgroup

\input{sections/introduction}
\input{sections/method}
\input{sections/experiments}
\input{sections/conclusion}
\input{sections/contributors}

\bibliography{references}
\bibliographystyle{references}

\clearpage 
\appendix
\input{sections/appendix}

\end{document}

%% file: sections/abstract.tex
\normalsize
Diffusion language models offer a promising alternative to autoregressive models due to their potential for parallel and iterative generation. However, existing approaches use a single network for both context representation and iterative denoising, forcing one model to serve both roles and limiting its capacity for either role. We propose \ourmethod, a block-wise autoregressive diffusion model that decouples these roles into two towers: a \emph{frozen} AR context tower that causally processes clean tokens, and a trainable diffusion denoiser tower with bidirectional block attention that refines noisy blocks via cross-attention to the context. Built on \backbone---an open-weight 30B hybrid Mamba-Transformer MoE model---and trained on ${\sim}2.1$T tokens, \ourmodel retains \textbf{98.7\%} of the autoregressive baseline's quality while offering \textbf{2.42$\times$} higher wall-clock generation throughput. We release the code and model weights in the \href{https://huggingface.co/collections/nvidia/nemotron-labs-twotower}{\ourmodel collection}.

%% file: sections/introduction.tex
\section{Introduction}
\label{sec:intro}

Autoregressive (AR) language models are the predominant paradigm in text generation~\citep{radford2019language, grattafiori2024llama, liu2024deepseek}. The one-token-at-a-time decoding process, however, creates a throughput bottleneck. Discrete diffusion language models take a different approach: they generate tokens in parallel and refine them iteratively, offering higher throughput and finer-grained controllability~\citep{austin2021structured, sahoo2024simple, nie2025large, arriola2025block}. In this paradigm, existing models typically use a single decoder for two distinct roles at every denoising step, representing the clean tokens and denoising the corrupted tokens. This entanglement pulls the same set of weights in different directions, limiting their capacity to excel at either.

\citet{arriola2025encoder} observed that these two roles can be handled by separate modules, proposing an encoder-decoder architecture where the encoder represents clean tokens and a lightweight decoder iteratively denoises each block. Their largest experiment, at 1.7B scale, trains both modules with tied weights. It remains to be seen whether fully decoupling the two roles holds at larger scales, particularly on modern hybrid architectures that combine Mamba, attention, and mixture-of-experts (MoE) layers.

We present \ourmethod and instantiate it as \ourmodel on \backbone~\citep{blakeman2025nvidia}, an open-weight hybrid model that interleaves Mamba-2, attention, and mixture-of-experts layers. We take two copies of the pretrained network and assign them complementary roles: the AR context tower is kept \emph{frozen} as a causal model over clean tokens, preserving the backbone's autoregressive capability; the diffusion denoiser tower is trained via a mask-diffusion objective with bidirectional self-attention within each noisy block and causal cross-attention to past clean context.

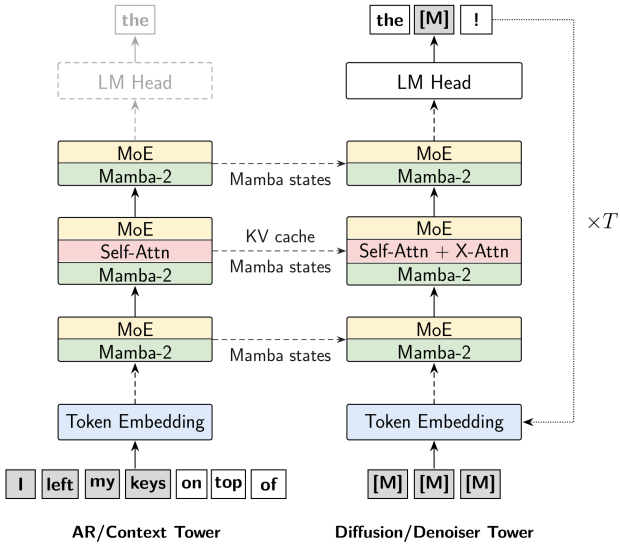
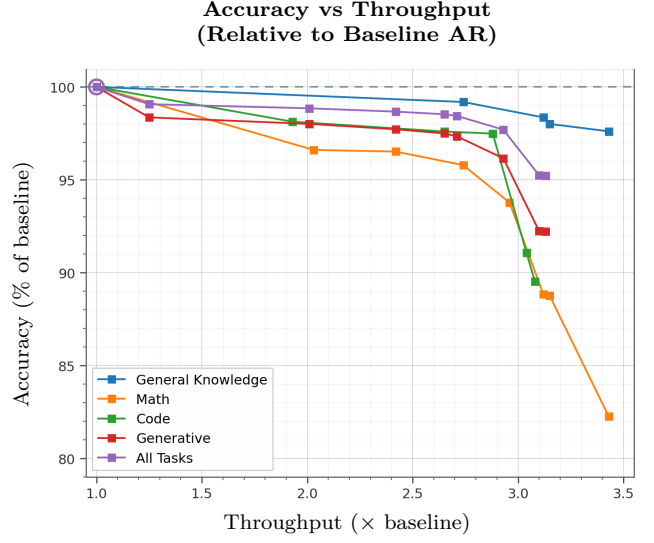
\begin{figure}[t]
    \centering
    \newlength{\overviewfigmaxht}
    \setlength{\overviewfigmaxht}{0.33\textheight}

    \begin{subfigure}[t]{0.49\linewidth}
        \centering
        \vspace{0pt}
        \begin{minipage}[t][\overviewfigmaxht][c]{\linewidth}
            \centering
            \includegraphics[width=\linewidth,height=\overviewfigmaxht,keepaspectratio]{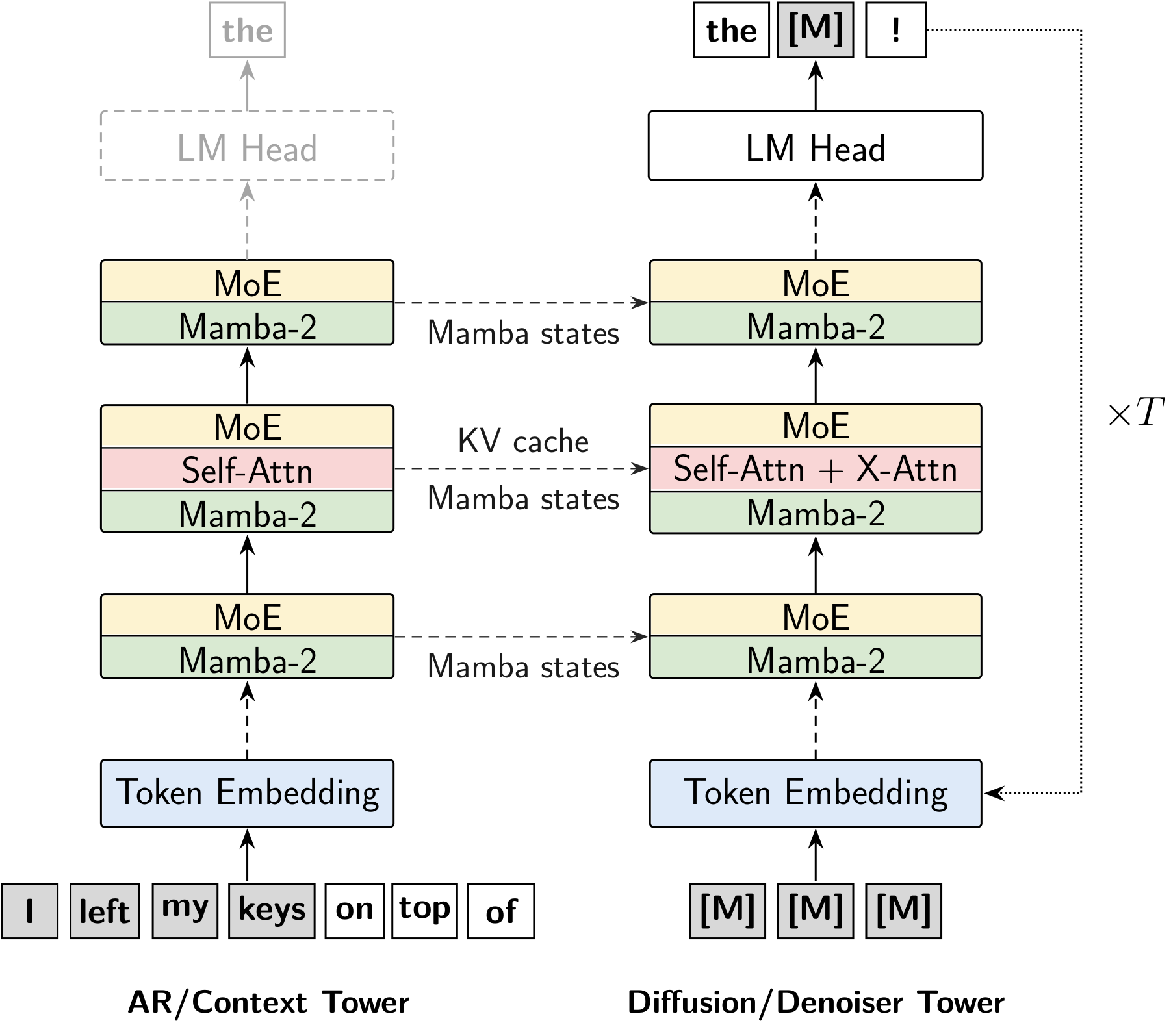}
        \end{minipage}
        \caption{Two-tower architecture: a frozen AR context tower conditions a diffusion denoiser tower over each noisy block.}
        \label{fig:twotower-arch}
    \end{subfigure}\hfill
    \begin{subfigure}[t]{0.49\linewidth}
        \centering
        \vspace{0pt}
        \begin{minipage}[t][\overviewfigmaxht][c]{\linewidth}
            \centering
            \begin{tikzpicture}
                \node[inner sep=0] (pareto) {\includegraphics[
                    width=\linewidth,
                    height=\overviewfigmaxht,
                    keepaspectratio,
                    trim=24 28 8 30,
                    clip
                ]{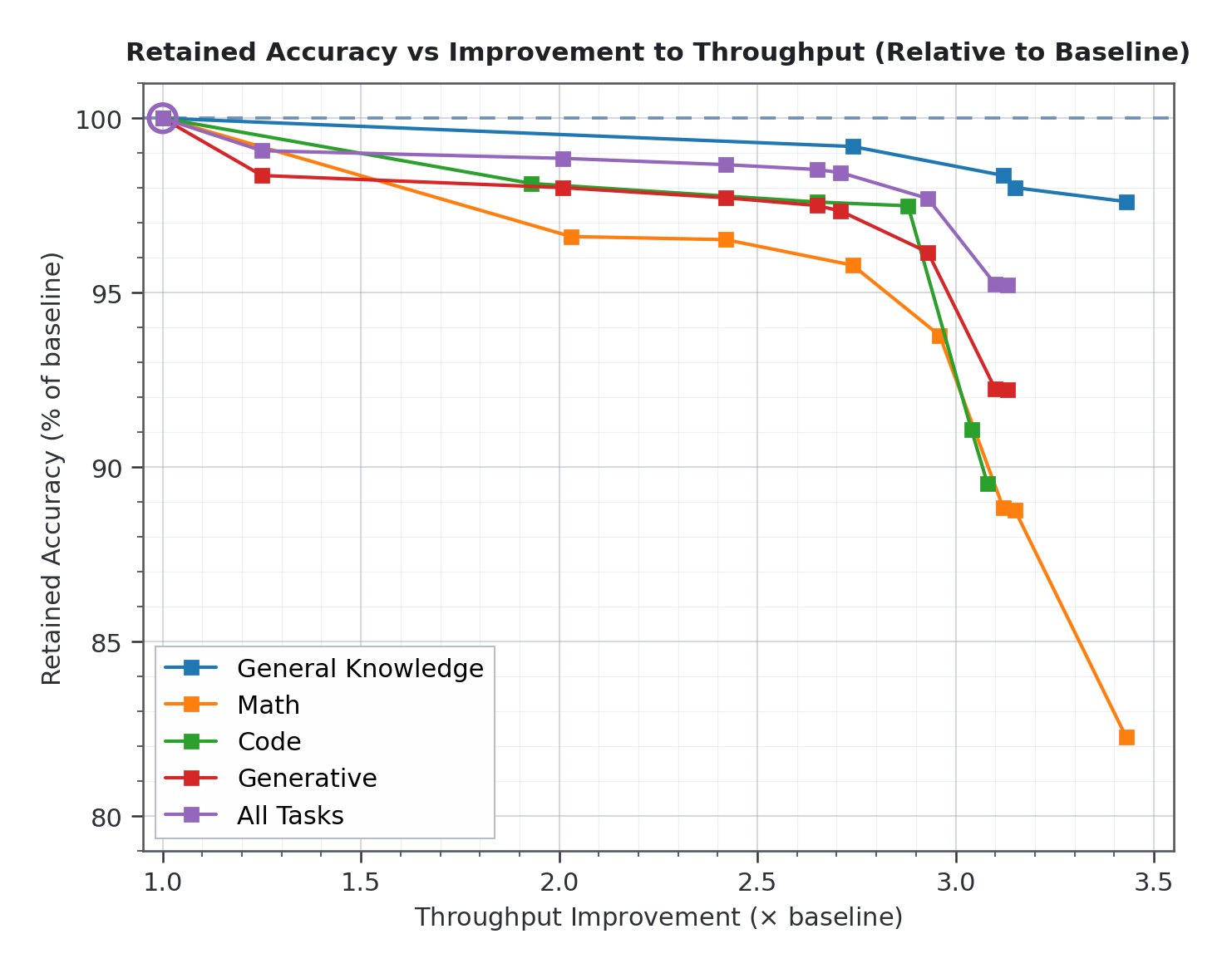}};
                \path[use as bounding box] ($(pareto.south west)+(-10pt,-14pt)$) rectangle ($(pareto.north east)+(10pt,18pt)$);
                \fill[white,overlay] ($(pareto.north west)+(-2pt,-3pt)$) rectangle ($(pareto.north east)+(2pt,19pt)$);
                \fill[white,overlay] ($(pareto.south west)+(-2pt,-13pt)$) rectangle ($(pareto.south east)+(2pt,3pt)$);
                \fill[white,overlay] ($(pareto.south west)+(-10pt,-2pt)$) rectangle ($(pareto.north west)+(4pt,2pt)$);
                \node[overlay,anchor=south, font=\scriptsize\bfseries, align=center, text width=0.92\linewidth, fill=white, inner sep=1pt]
                    at ($(pareto.north)+(0,4pt)$)
                    {Accuracy vs Throughput\\(Relative to Baseline AR)};
                \node[overlay,anchor=north, font=\scriptsize, fill=white, inner sep=1pt]
                    at ($(pareto.south)+(0,-1pt)$)
                    {Throughput ($\times$ baseline)};
                \node[overlay,anchor=center, rotate=90, font=\scriptsize, fill=white, inner sep=1pt]
                    at ($(pareto.west)+(-6pt,0)$)
                    {Accuracy (\% of baseline)};
            \end{tikzpicture}
        \end{minipage}
        \caption{Quality--throughput curve across confidence thresholds and denoising budgets, normalized to the AR baseline.}
        \label{fig:pareto}
    \end{subfigure}

    \caption{Overview of \ourmodel. (a) The frozen AR context tower runs causally over the prompt and committed tokens, preserving the pretrained backbone and maintaining reusable KV cache and Mamba states. The trainable diffusion denoiser tower iteratively resolves each noisy block using layer-aligned context attention and context-seeded Mamba states. (b) Quality--throughput trade-off relative to the one-token-at-a-time AR baseline.}
    \label{fig:overview}
\end{figure}

The towers connect layer-by-layer. Each denoiser layer cross-attends to the corresponding layer of the context tower, giving the denoiser multi-scale access to the backbone's representations. This is in contrast to prior approaches that broadcast only the last hidden state. The denoiser is further modulated by the diffusion timestep via adaptive layer normalization~\citep{peebles2023scalable}, which is common in image diffusion transformers but less so in masked diffusion language models~\citep{ou2024your}.

Trained on ${\sim}2.1$T tokens---a fraction of the total 25T tokens used to pretrain the backbone---\ourmodel preserves most of the autoregressive baseline's quality while delivering substantially higher throughput on a 30B hybrid MoE architecture. At the default operating point, \ourmodel commits multiple tokens per refinement step early in decoding, helping explain the observed wall-clock speedup over one-token-at-a-time AR decoding. We release the training recipe, code, and model weights in the \href{https://huggingface.co/collections/nvidia/nemotron-labs-twotower}{\ourmodel collection}.

%% file: sections/method.tex
\section{Method}
\label{sec:method}

In this section we describe the two-tower architecture (Section~\ref{subsec:arch}), the masked diffusion formulation (Section~\ref{subsec:diffusion}), and the training recipe (Section~\ref{subsec:training}).

\subsection{Two-Tower Architecture}
\label{subsec:arch}

\ourmethod is a general approach that can be applied to any pretrained autoregressive language model. In this work we instantiate it on \backbone~\citep{blakeman2025nvidia}, which consists of 52 layers: 23 Mamba-2 layers, 6 self-attention layers, and 23 mixture-of-experts (MoE) layers.

We create two copies of this network and assign them complementary roles (see Figure~\ref{fig:twotower-arch}). Given a prompt, the \emph{AR context tower} acts as a causal autoregressive model over the prompt and previously committed tokens, producing per-layer KV pairs and final Mamba states. Generation then proceeds block by block: each new block is initialized with $S$ \mask tokens and iteratively refined by the \emph{diffusion denoiser tower} over $T$ denoising steps. At every layer of the denoiser, attention layers cross-attend to the corresponding context-tower KV cache, while Mamba-2 layers seed their initial state from the corresponding context-tower Mamba state. Once all tokens in a block are clean, the block is committed; the context tower processes the committed tokens causally to update its caches, and generation continues with the next block. Algorithm~\ref{alg:twotower-gen} outlines this block-wise autoregressive generation process.

\begin{algorithm}[t]
\caption{\ourmethod block-wise generation.}
\label{alg:twotower-gen}
\begin{algorithmic}[1]
\REQUIRE Prompt $\mathbf{x}_\text{prompt}$, block size $S$, denoising steps $T$
\STATE $(\text{KV}, \text{States}) \gets \text{ContextTower}(\mathbf{x}_\text{prompt})$ \hfill $\triangleright$ Build AR context
\FOR{each block $b = 1, 2, \dots$}
    \STATE $\mathbf{z}^b \gets (\mask,\dots,\mask)$
    \FOR{$t = 1,\dots,T$}
        \STATE $\mathbf{z}^b \gets \textsc{SampleStep}\bigl(\mathbf{z}^b;\, \text{DenoiserTower}(\mathbf{z}^b, t, \text{KV}, \text{States})\bigr)$
    \ENDFOR
    \STATE $\mathbf{x}_b \gets \mathbf{z}^b$ \hfill $\triangleright$ Commit block
    \STATE $(\text{KV}, \text{States}) \gets \text{ContextTower}(\mathbf{x}_b, \text{KV}, \text{States})$ \hfill $\triangleright$ Update AR context
\ENDFOR
\end{algorithmic}
\end{algorithm}

At inference time, \ourmethod introduces a fixed memory footprint for the context tower weights, while keeping a single prefix cache. Only the context tower maintains KV and Mamba states across blocks, updating them as blocks are committed, so the sequence-length-dependent cache memory scales like the AR baseline.

We keep the context tower body unchanged. Its final vocabulary projection/LM head is optional: it can be omitted in the default diffusion-generation path, where the context tower only needs to produce states, and retained when the context tower is used for speculative decoding,verification, likelihood evaluation, or AR scoring. In contrast, we make a few architectural modifications to the denoiser tower to adapt it for diffusion training and inference. We describe each modification below, and provide ablations in Section~\ref{subsec:exp-ablations}.

\newparagraph{Bidirectional Attention.} Within the block under refinement, the denoiser relaxes the causal mask: noisy tokens attend bidirectionally to other noisy tokens while remaining causal with respect to past clean blocks. This adds no parameters and, for standard dense attention kernels, does not change per-layer FLOPs. At denoiser layer $i$, queries from the noisy block attend over a concatenated key/value sequence of the context tower's layer-$i$ KV for past blocks $0,\dots,b{-}1$ and the denoiser's own layer-$i$ KV for block $b$:
\begin{equation}
    \text{Attn}\big(\mathbf{Q}_b^{(i)},\; [\mathbf{K}_{<b}^{\text{ctx},(i)};\, \mathbf{K}_b^{\text{den},(i)}],\; [\mathbf{V}_{<b}^{\text{ctx},(i)};\, \mathbf{V}_b^{\text{den},(i)}]\big).
    \label{eq:xattn}
\end{equation}
The cross-attention is \emph{layer-aligned}: denoiser layer $i$ attends to context layer $i$. Because both towers initialize from the same pretrained checkpoint, same-index layers operate at comparable representation levels, making layer-aligned cross-attention a natural pairing.

\newparagraph{Bidirectional Mamba.} We also test a parameter-free bidirectional Mamba-2 variant by running the pretrained Mamba-2 weights left-to-right and right-to-left from zero states, then averaging the outputs. This roughly doubles per-layer state-space model (SSM) FLOPs; ablations show only marginal quality gains, so the final design keeps Mamba causal.

\newparagraph{Time Conditioning.} We condition the denoiser on the timestep $t$ using adaLN-single~\citep{peebles2023scalable, chen2023pixart}. A global MLP maps $t$ to shared scale, shift, and gate parameters, with per-layer learned embeddings for layer-specific modulation. On the 30B backbone, this adds only 1.5M parameters; we replicate these small modules on each tensor-parallel rank rather than sharding them, avoiding extra communication cost. Time conditioning improves sampling quality (Section~\ref{subsec:exp-ablations}).

\newparagraph{Expert Routing.} We use the backbone's existing MoE routing mechanism, including sequence-level load balancing. Tokens are noise-aware through adaLN modulation. We make no explicit routing choices and allow experts to learn specialization from the denoising objective.

\subsection{Block Diffusion Language Modeling}
\label{subsec:diffusion}

We train the denoiser under the masked diffusion framework~\citep{sahoo2024simple, shi2024simplified, ou2024your}, applied to each autoregressive block. Let $\mathbf{x}_b=(x_b^1,\dots,x_b^S)$ denote the $b$-th block of $S$ tokens, and let $\mathbf{c}_{<b}=(\text{KV}_{<b},\text{States}_{b-1})$ denote the context-tower attention KV cache over previous clean blocks and the per-layer Mamba boundary states after block $b{-}1$. We model the sequence distribution as block-autoregressive:
\begin{equation}
    \log p_\theta(\mathbf{x}) = \sum_{b=1}^{B} \log p_\theta(\mathbf{x}_b \mid \mathbf{x}_{<b}),
    \label{eq:md-factorization}
\end{equation}
where each conditional block distribution is parameterized by a masked diffusion process conditioned on $\mathbf{c}_{<b}$.

For diffusion time $t \in (0,1]$, the forward process corrupts a clean block $\mathbf{x}_b$ into a noisy block $\mathbf{z}_t^b$ by independently replacing each token with \mask. The schedule $\alpha_t$ is the probability that a clean token remains unmasked at noise level $t$, so $1-\alpha_t$ is the masking probability; in our experiments we use the linear schedule $\alpha_t = 1-t$. Using $\delta_y$ to denote a point mass at token $y$ and $\mathbf{m}$ for the \mask point mass, the per-token marginal is
\begin{equation}
    q(\mathbf{z}_t^{b,\ell} \mid x_b^\ell)
    =
    \text{Cat}\!\left(
    \mathbf{z}_t^{b,\ell};\,
    \alpha_t \delta_{x_b^\ell}
    + (1-\alpha_t)\mathbf{m}
    \right),
    \label{eq:md-forward}
\end{equation}
where $\alpha_t$ decreases from nearly $1$ to $0$ as $t$ increases. Small $t$ corresponds to lightly corrupted blocks, while large $t$ approaches a fully masked block.

The denoiser predicts the clean token at masked positions, conditioned on the noisy block, the diffusion time, and the context cache. The masked diffusion ELBO motivates a time-weighted negative log-likelihood over masked tokens~\citep{sahoo2024simple, shi2024simplified}; for the linear schedule $\alpha_t=1-t$, the theoretical weight simplifies to $1/t$. In training, we omit this importance weight for stability and optimize the mean negative log-likelihood over masked positions:
\begin{equation}
    \mathcal{L}_\mathrm{MD} =
    \mathbb{E}_{t, \mathbf{z}_t}
    \left[
    \frac{1}{|\mathcal{M}_t|}
    \sum_{(b,\ell)\in\mathcal{M}_t}
    -\log p_\theta\!\left(
    x_b^\ell \mid \mathbf{z}_t^b, t, \mathbf{c}_{<b}
    \right)
    \right].
    \label{eq:md-loss}
\end{equation}
Here $\mathcal{M}_t=\{(b,\ell): z_t^{b,\ell}=\mask\}$ is the set of masked token positions.

\newparagraph{Sampling.} Algorithm~\ref{alg:twotower-gen} describes the outer block-autoregressive loop, while Algorithm~\ref{alg:confidence-unmasking} summarizes the single-block sampler used in our main results. The sampler receives the prefix caches $(\text{KV}_{<b},\text{States}_{b-1})$ from the context tower and keeps them fixed while refining block $b$; the caches are advanced only after the block is committed. We also consider the standard predict-and-noise sampler used in block diffusion models~\citep{arriola2025block, arriola2025encoder}, which predicts a clean block and then re-masks low-confidence positions according to the noise schedule. Our main results use a confidence-unmasking variant with threshold $\gamma$: at each step, the denoiser predicts all currently masked tokens in parallel, predictions above $\gamma$ are committed, and the remaining uncertain positions stay masked for later refinement. This makes the number of committed tokens adaptive to the model's confidence while ensuring the block is completed within $T$ steps.

\begin{algorithm}[t]
\caption{Single-block confidence unmasking.}
\label{alg:confidence-unmasking}
\begin{algorithmic}[1]
\REQUIRE Prefix caches $(\text{KV}_{<b}, \text{States}_{b-1})$, block size $S$, steps $T$, confidence threshold $\gamma$
\STATE $\mathbf{z} \gets (\mask,\dots,\mask)$
\FOR{$r=1,\dots,T$}
    \STATE $\tau \gets |\{\ell: z^\ell=\mask\}|/S$
    \STATE $p_\theta \gets \text{DenoiserTower}(\mathbf{z}, \tau, \text{KV}_{<b}, \text{States}_{b-1})$
    \STATE $\mathbf{z} \gets \textsc{ConfidenceSample}(\mathbf{z}, p_\theta, \gamma, T-r+1)$
\ENDFOR
\STATE \textbf{return} $\mathbf{z}$
\end{algorithmic}
\end{algorithm}

\subsection{Training}
\label{subsec:training}

We train only the denoiser tower, while using the context tower as a frozen AR representation model. Training follows the two-stage curriculum of the \backboneshort backbone: an initial broad-coverage phase followed by a higher-quality refinement phase.

\newparagraph{Initialization and data.} Phase~1 initializes both towers from the same pretrained \backbone base checkpoint and trains the denoiser on a subset of the \backboneshort phase-1 blend, which emphasizes data diversity and broad coverage. Phase~2 continues the denoiser from Phase~1, and switches to the \backboneshort phase-2 blend, which upweights higher-quality and STEM-focused sources. Across both phases, we train on roughly ${\sim}1.4$T tokens.

\newparagraph{Optimization.} We use BF16 precision, AdamW \citep{loshchilov2019decoupledweightdecayregularization}, and a Warmup-Stable-Decay learning rate schedule \citep{hu2024minicpmunveilingpotentialsmall} with peak learning rate \num{1e-4} and final learning rate \num{1e-6}. The optimizer and learning-rate schedule are reset at phase boundaries. Our final configuration uses block size $S{=}16$; in the current implementation, the Mamba chunk size is matched to $S$ so the existing Mamba kernel exposes states at diffusion block boundaries.

\newparagraph{Implementation.} We implemented the training loop in Megatron-LM\footnote{\url{https://github.com/nvidia/megatron-lm}} \citep{megatron-lm}. For each clean sequence, the frozen context tower runs once with no-grad under its standard causal AR mask. Its attention layers produce layer-aligned KV caches, while its Mamba-2 layers expose the recurrent conv and SSM states after every clean block. These block-boundary states are the states a causal Mamba layer would carry forward after consuming each block. The denoiser starts from the same clean sequence, applies the mask-diffusion corruption, and processes all noisy blocks in one forward pass: attention layers attend to past clean context blocks and bidirectionally within the current noisy block, while Mamba layers fold blocks into the batch dimension. Block~0 starts from a zero Mamba state; block~$b$ starts from the context-tower Mamba state after block~$b{-}1$. This batched formulation exposes the denoiser to diverse block-level corruption patterns within each optimizer step, without increasing the underlying sequence batch.

%% file: sections/experiments.tex
\section{Results}
\label{sec:experiments}

We evaluate \ourmodel against the autoregressive \backbone baseline used to initialize both towers. We report base-model benchmarks spanning general knowledge, code, math, commonsense, and multilingual tasks, using checkpoints after long-context pretraining and before instruction tuning, RL, or alignment. All evaluations use BF16 precision on 2$\times$H100 GPUs. Throughput is measured from the wall-clock time to produce the final answer under the same inference setup, and reported as speedup relative to the AR baseline.

\subsection{Quality--Throughput Trade-off}
\label{subsec:exp-main}

\begin{figure}[t]
\centering
\includegraphics[width=\linewidth]{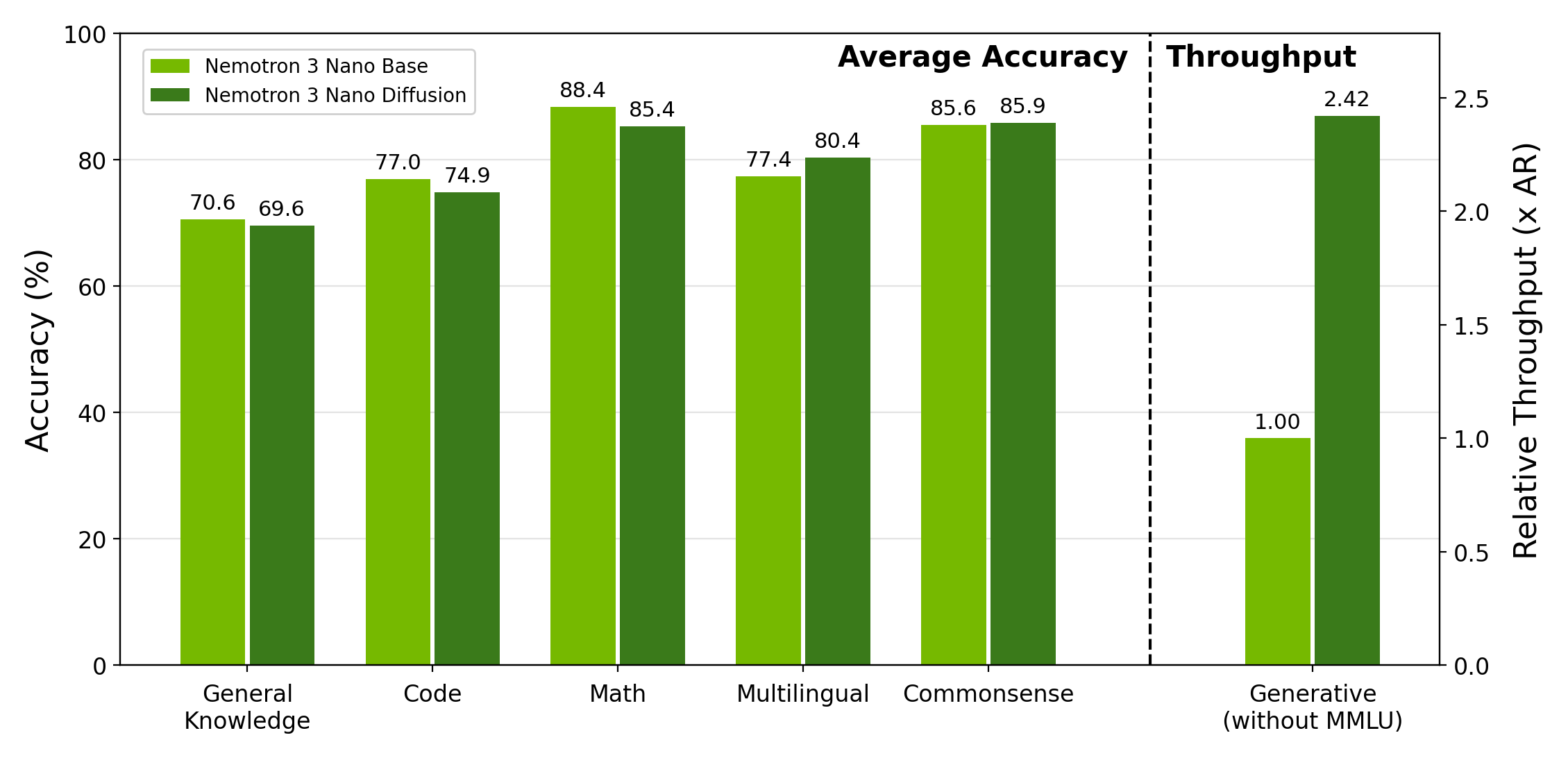}
\caption{Category-level comparison between the \backbone autoregressive baseline and \ourmodel. Left: average accuracy by benchmark category. Right: relative wall-clock throughput for generative evaluations. \ourmodel preserves most of the AR baseline's quality while improving relative generation throughput to $\mathbf{2.42\times}$.}
\label{fig:bar-results}
\end{figure}

\ourmodel preserves \textbf{98.7\%} of the AR baseline's aggregate benchmark quality while improving wall-clock generation throughput by \textbf{2.42$\times$}. This operating point lies on the quality--throughput curve in Figure~\ref{fig:pareto}; Figure~\ref{fig:bar-results} expands the same point into category-level accuracy and relative throughput. The result shows that pretrained AR representations can support a trainable diffusion denoiser while preserving broad benchmark quality and yielding faster generation. We use confidence unmasking (Algorithm~\ref{alg:confidence-unmasking}) with threshold $\gamma{=}0.8$ and block size $S{=}16$. Varying the confidence threshold $\gamma$ and number of denoising steps $T$ traces the broader curve. Higher-confidence settings recover slightly more quality, while lower thresholds accept more tokens per step and reach throughput beyond $3\times$ with larger quality loss. We report the $\gamma{=}0.8$ point as the best balance we found between preserving the backbone's capability and realizing decoding-speed gains.

The benchmark task breakdown is shown in Figure~\ref{fig:bar-results}. General knowledge remains within about one point of the AR baseline, code and math show modest degradation, and commonsense and multilingual performance are recovered or improved. The Pareto curve and task breakdown indicate that the two-tower diffusion model preserves most of the pretrained model's behavior. Notably, this quality recovery comes from adaptation rather than re-pretraining. \ourmodel starts from an off-the-shelf AR checkpoint and trains the denoiser on only ${\sim}2.1$T tokens, a fraction of the 25T tokens used to pretrain the backbone.

\subsection{Design Choices}
\label{subsec:exp-ablations}

Table~\ref{tab:design-progression} tracks the main design choices on an internal \backbone base checkpoint. The ablations compare bidirectional attention with causal Mamba, bidirectional Mamba, adaLN time conditioning, and continued training on the phase-2 blend. We also study whether the pretrained AR representation should remain fixed during denoiser adaptation (Table~\ref{tab:context-tower-training}).

\begin{table}[t]
\centering
\small
\caption{Design choices for \ourmodel using block size $S{=}32$ on an internal \backbone base checkpoint. We start with bidirectional attention, causal Mamba, and the phase-1 data blend, then add bidirectional Mamba, adaLN time conditioning, and continued training on the phase-2 data blend.}
\label{tab:design-progression}
\begin{tabular}{@{}lcc ccc@{}}
\toprule
   &        &      & \multicolumn{3}{c}{Accuracy (\%)} \\
\cmidrule(lr){4-6}
Config & \shortstack{Causal\\ Mamba} & Blend & Gen. & Code & Math \\
\midrule
Bidirectional Attention              & \checkmark  & phase1 & 72.94 & 68.64 & 80.57 \\
+~Bidirectional Mamba                & $\times$    & phase1 & 72.96 & 68.05 & 79.78 \\
+~Time Conditioning                  & \checkmark  & phase1 & 74.12 & 69.61 & 81.30 \\
+~Phase~2 Stage                      & \checkmark  & phase2 & \textbf{75.11} & \textbf{71.51} & \textbf{82.08} \\
\bottomrule
\end{tabular}
\end{table}

\newparagraph{Bidirectional Mamba.} A right-to-left Mamba scan leaves generation essentially unchanged (72.94 to 72.96) while reducing code and math (68.64 to 68.05 and 80.57 to 79.78). Since this variant roughly doubles denoiser SSM compute, we keep Mamba causal and rely on bidirectional attention within the noisy block for denoising context.

\newparagraph{Time conditioning.} AdaLN time conditioning improves generation, code, and math from 72.94, 68.64, and 80.57 to 74.12, 69.61, and 81.30, with only 1.5M added parameters. The gain shows that the denoiser benefits from explicit access to the current noise level.

\newparagraph{Data curriculum.} Continuing the time-conditioned denoiser on the phase-2 blend further improves generation, code, and math to 75.11/71.51/82.08. This mirrors the backbone's own curriculum. Broad phase-1 coverage is useful for initial adaptation, while the higher-quality phase-2 blend improves downstream generation, code, and math.

\newparagraph{Tower Decoupling.} The decoupled two-tower design gives the smallest quality drop, while continued backbone AR training leads to lower accuracy on generation and math. Tying the towers, i.e., sharing weights between the towers, under a joint AR+diffusion loss is substantially worse under both AR and diffusion decoding modes. These results suggest keeping the context tower frozen and adapting a separate denoiser.

\begin{table}[t]
\centering
\footnotesize
\caption{Tower-decoupling ablation after $\sim167$B phase-1 tokens, reported as relative accuracy change from \backbone. \ourmodel keeps the context tower frozen and trains a separate denoiser. Alternative rows continue backbone AR training or tie the two towers under a joint AR+diffusion objective, evaluated with AR or diffusion decoding as indicated.}
\label{tab:context-tower-training}
\begin{tabular}{@{}lccccccc@{}}
\toprule
 & \multicolumn{2}{c}{Training Updates} & & \multicolumn{1}{c}{Inference} & \multicolumn{3}{c}{Relative Accuracy (\%)} \\
\cmidrule(lr){2-3}\cmidrule(lr){5-5}\cmidrule(lr){6-8}
Config & Context & Denoiser & \shortstack{Weights\\Shared} & \shortstack{Decoding\\Mode} & Gen. & Code & Math \\
\midrule
\backbone{} & -- & -- & n/a & AR & +0.0 & +0.0 & +0.0 \\
\quad$\hookrightarrow$ Continued AR training & \checkmark & -- & n/a & AR & -10.5 & \textbf{-8.3} & -17.8 \\
\ourmodel{} & $\times$ & \checkmark & $\times$ & Diffusion & \textbf{-6.2} & -10.5 & \textbf{-11.3} \\
\quad$\hookrightarrow$ Joint loss, tied towers & \checkmark & \checkmark & \checkmark & AR & -26.2 & -21.0 & -26.4 \\
\quad$\hookrightarrow$ Joint loss, tied towers & \checkmark & \checkmark & \checkmark & Diffusion & -27.9 & -27.8 & -27.0 \\
\bottomrule
\end{tabular}
\end{table}

\subsection{Training Block Size}
\label{subsec:exp-block}

Training block size controls how many tokens the denoiser resolves before the context tower is updated. Larger blocks expose more parallelism but require denoising a longer span from a fixed prefix; smaller blocks stay closer to autoregressive conditioning at lower speedup. Table~\ref{tab:block} reports block-size ablations on the internal \backbone checkpoint.

\begin{table}[t]
\centering
\small
\caption{Training block-size ablations on an internal \backbone checkpoint. The first three rows sweep phase-1 training block size; the next three rows continue training on the phase-2 blend. Training and inference use the same block size. Throughput is reported as wall-clock speedup over the AR baseline for each ablation run.}
\label{tab:block}
\begin{tabular}{@{}cc ccc c@{}}
\toprule
           &      & \multicolumn{3}{c}{Accuracy (\%)} & \\
\cmidrule(lr){3-5}
Block size & Blend & Gen. & Code & Math & Throughput ($\times$ AR) \\
\midrule
128 & phase1 & 69.59 & 63.56 & 75.73 & 2.23 \\
64  & phase1 & 71.97 & 66.61 & 79.47 & 2.28 \\
32  & phase1 & 72.94 & 68.64 & 80.57 & 2.17 \\
\midrule
32  & phase2 & 76.36 & 73.84 & 82.24 & 2.25 \\
16  & phase2 & 77.10 & 74.56 & 85.45 & 2.02 \\
 8  & phase2 & 77.23 & 73.79 & 85.98 & 1.71 \\
\bottomrule
\end{tabular}
\end{table}

The trend is consistent. Reducing the training block size improves quality. In phase 1, moving from $S{=}128$ to $S{=}32$ improves generation, code, and math from 69.59/63.56/75.73 to 72.94/68.64/80.57. After phase-2 continuation, $S{=}16$ reaches 77.10/74.56/85.45, while $S{=}8$ gives only marginal additional generation and math quality at lower throughput. We use $S{=}16$ as the default training block size because it captures most of the quality gain from smaller blocks while retaining a clear speed advantage over the AR baseline.

\subsection{Sampling Block Size}
\label{subsec:exp-iblock}

We next separate training block size from sampling block size. Table~\ref{tab:iblock} fixes the phase-2 checkpoint trained at $S{=}16$ and varies only the block size used during sampling.

\begin{table}[t]
\centering
\small
\caption{Sampling block-size sensitivity. We fix the \ourmodel checkpoint trained at block size $S{=}16$ on the phase-2 blend and vary only the sampling block size.}
\label{tab:iblock}
\begin{tabular}{@{}c ccccc@{}}
\toprule
           & \multicolumn{5}{c}{Accuracy (\%)} \\
\cmidrule(lr){2-6}
Block size & MMLU & HumanEval & GSM8K & MATH-500 & Multilingual \\
\midrule
64 & 78.10 &          19.85  &           2.20  &           2.20  &          37.75  \\
32 & 78.19 &          33.60  &          61.71  &          46.95  &          65.22  \\
16 & 78.32 & \textbf{76.40}  &          89.84  &          81.05  &          77.15  \\
 8 & \textbf{78.37} & 73.14  & \textbf{90.30}  & \textbf{81.65}  & \textbf{77.58}  \\
\bottomrule
\end{tabular}
\end{table}

The results are asymmetric. Sampling with blocks larger than the training block size is poorly matched to the denoiser, especially on generation-heavy tasks. HumanEval falls from 76.40 at $S{=}16$ to 19.85 at $S{=}64$, while GSM8K and MATH-500 fall to 2.20. Sampling with smaller blocks is more robust: the $S{=}8$ setting slightly improves MMLU, GSM8K, MATH-500, and multilingual accuracy relative to $S{=}16$. However, smaller blocks require more frequent context-tower updates, which lowers throughput as shown in Table~\ref{tab:block}. We therefore use $S{=}16$ as the default block size, since it preserves most of the quality benefit of smaller blocks while retaining higher throughput.

\subsection{Released Checkpoint}
\label{subsec:released-checkpoint}

For the released \ourmodel checkpoint, we initialize from the final \backbone base model and train the denoiser in three stages: phase-1 adaptation at block size $S{=}32$, phase-2 continuation at $S{=}32$, and a final phase-2 continuation at $S{=}16$. The first two stages use the larger block size for training efficiency, while the last stage adapts the denoiser to the default block size used for sampling. The quality--throughput curve in Figure~\ref{fig:pareto} and the category-level results in Figure~\ref{fig:bar-results} are reported from this released checkpoint.

\subsection{Sampling Dynamics}
\label{subsec:exp-sampling-dynamics}

\begin{figure*}[t]
\centering
\includegraphics[width=\linewidth]{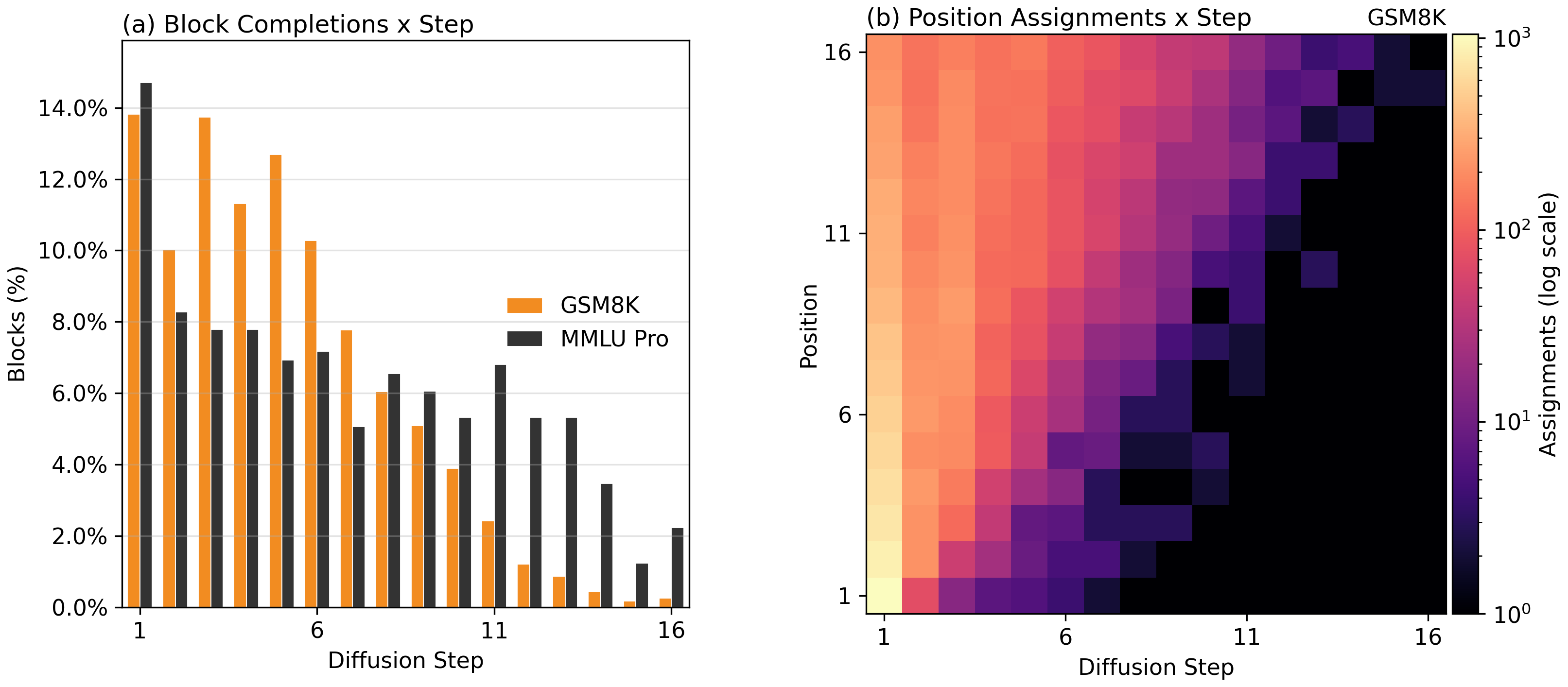}
\caption{%
Sampling dynamics for the released \ourmodel checkpoint ($\gamma{=}0.8$, $S{=}16$). Statistics are computed from generation traces over 100 problems per task and include only blocks that contribute to the extracted answer; Appendix~\ref{sec:appendix-sampling-dynamics} shows the full task set and unfiltered traces.
\textbf{(a)} Percentage of answer-producing blocks that complete at each diffusion step for GSM8K and MMLU-Pro. Most blocks finish in the first few steps, with MMLU-Pro showing a broader tail.
\textbf{(b)} Distribution of token commitments by position and diffusion step for GSM8K (log scale; zero counts are mapped to one for visualization). Earlier positions are usually committed earlier, producing an autoregressive upper-left triangular pattern.
}
\label{fig:sampling-dynamics}
\end{figure*}

We study the sampling behavior of the released \ourmodel checkpoint under the default confidence-unmasking setup ($\gamma{=}0.8$, $S{=}16$). Here a block denotes the group of $S$ tokens refined jointly before being committed to the context tower. We sample 100 problems from each multi-token generative benchmark and record the diffusion trace for every answer-producing block; Figure~\ref{fig:sampling-dynamics} shows representative task views, and Appendix~\ref{sec:appendix-sampling-dynamics} extends the analysis across the full task set.

Block completion is adaptive. In Figure~\ref{fig:sampling-dynamics}a, both GSM8K and MMLU-Pro complete many blocks in the first few diffusion steps, but MMLU-Pro retains more mass at intermediate and late steps. Appendix~\ref{sec:appendix-sampling-dynamics} shows the same pattern across tasks: MGSM and MBPP-Sanitized often finish blocks immediately (Figure~\ref{fig:appendix-block-x-step}), while code and reasoning tasks sustain longer tails.

\begin{figure}[t]
\centering
\includegraphics[width=0.95\linewidth]{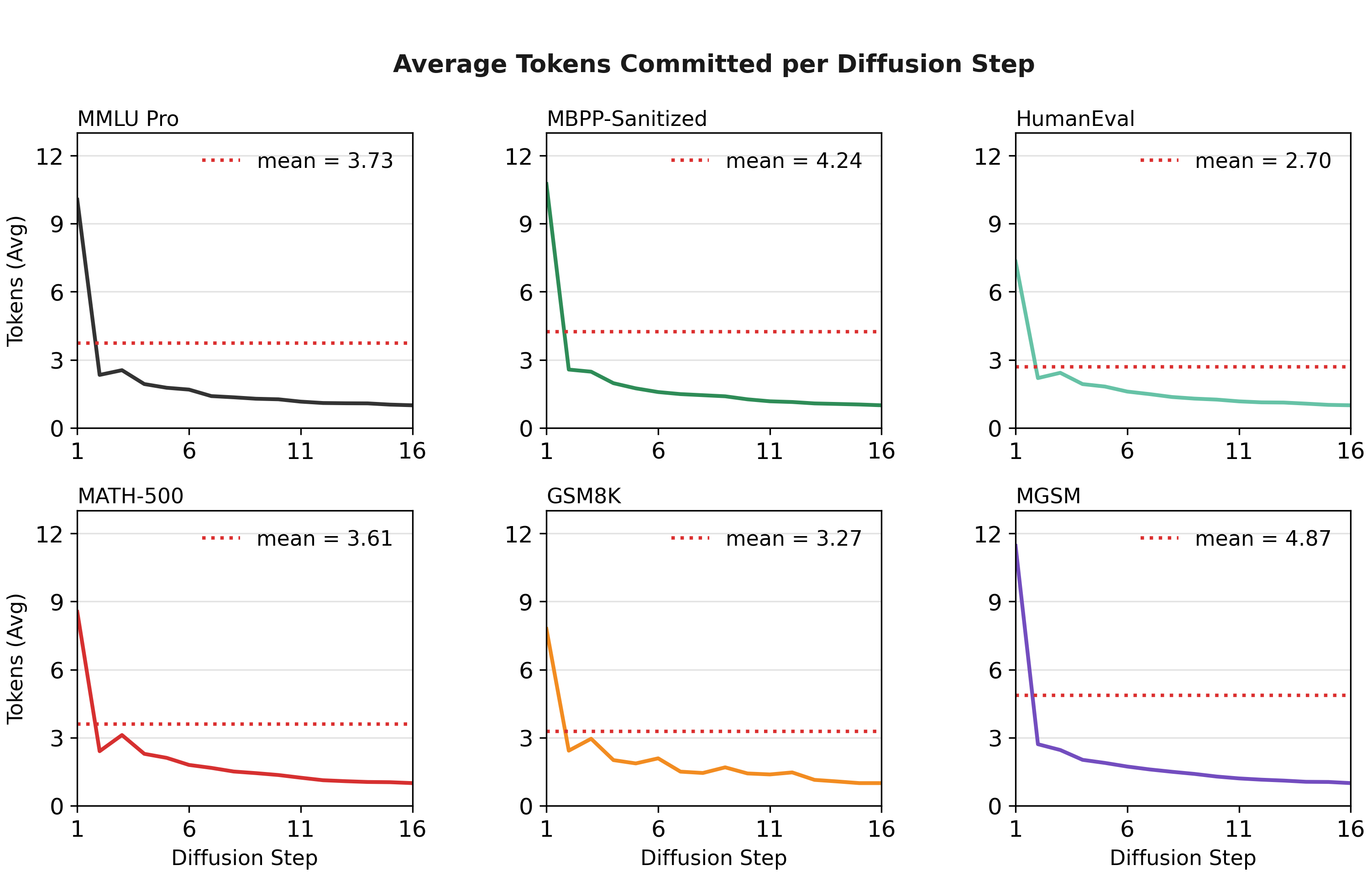}
\caption{Average committed tokens per diffusion step. The first diffusion step commits the most tokens, after which the count drops as the sampler focuses on the remaining low-confidence positions.}
\label{fig:sampling-dynamics-token-x-step}
\end{figure}

Commitment counts are strongly front-loaded in time. Figure~\ref{fig:sampling-dynamics-token-x-step} shows that the first diffusion step commits the largest number of tokens on average, after which the count drops quickly. Autoregressive decoding commits exactly one token per step; in contrast, \ourmodel commits multiple tokens per step early in refinement, helping explain how iterative block refinement can still yield wall-clock gains despite taking multiple denoising steps. This behavior is consistent with Algorithm~\ref{alg:confidence-unmasking}: each step predicts all masked positions in parallel, commits the high-confidence subset, and leaves the harder residual positions for later refinement.

Within a block, commitments also follow a left-to-right ordering. Figure~\ref{fig:sampling-dynamics}b shows that earlier positions are typically committed earlier, while later positions are more likely to survive to later diffusion steps, producing an upper-left triangular pattern. Appendix~\ref{sec:appendix-sampling-dynamics} shows that this behavior is not unique to GSM8K and appears across tasks with varying strength (Figure~\ref{fig:appendix-position-x-step}). A plausible explanation is a strong left-to-right inductive bias inherited from the pretrained backbone: the context tower is fully causal, and the denoiser combines bidirectional block attention with the same unidirectional Mamba structure; since the backbone contains 23 Mamba layers and only 6 attention layers, this bias may dominate the resulting sampling behavior.

%% file: sections/conclusion.tex
\section{Conclusion}
\label{sec:conclusion}

We presented \ourmethod, a two-tower diffusion language modeling approach that adapts a pretrained autoregressive backbone into a block-wise diffusion generator. The context tower preserves the backbone's causal representation; the denoiser refines noisy blocks with bidirectional block attention, layer-aligned context attention, and context-seeded Mamba states. This separation lets the model reuse the structure learned during autoregressive pretraining while converting token-by-token decoding into iterative block refinement.

On the 30B hybrid backbone, \ourmodel preserves most of the AR baseline's quality while delivering higher wall-clock generation throughput. The ablations show bidirectional attention and time conditioning improve denoising, causal Mamba is preferable to bidirectional Mamba, and block size/confidence control the quality--throughput trade-off. The released checkpoint starts from an off-the-shelf AR model and trains the denoiser on a fraction of the backbone pretraining budget. At inference time, \ourmodel keeps the context-tower weights resident alongside the diffusion denoiser, increasing the fixed model-weight memory footprint while maintaining a single persistent prefix cache, so the sequence-length-dependent cache memory scales like the AR baseline.

The results show that masked diffusion can serve as a practical decoding adaptation for large pretrained AR models, including hybrid MoE backbones. We release weights, training code, and recipe in the \href{https://huggingface.co/collections/nvidia/nemotron-labs-twotower}{\ourmodel collection}. In future updates, we plan to add post-trained \ourmodel models to the same collection.

%% file: sections/contributors.tex
\section*{Acknowledgments}

We thank Jared Casper, Yonggan Fu, Abhinav Garg, Jiantao Jiao, Ante Jukic, Mikail Khona, Markus Kliegl, Karsten Kreis, Pavlo Molchanov, Rauf Nasretdinov, Keshav Santhanam, Kevin Shih, and David Tarjan for helpful discussions.

%% file: sections/appendix.tex
\section{Additional Sampling Figures}
\label{sec:appendix-sampling-dynamics}

Unlike Figure~\ref{fig:sampling-dynamics}, these visualizations (and Figure~\ref{fig:sampling-dynamics-token-x-step}) include all generated blocks from Section~\ref{subsec:exp-sampling-dynamics}, including those beyond the extracted answer. They provide a complete view of the sampler, and are not restricted to answer-producing blocks alone.

\begin{figure}[H]
  \centering
  \includegraphics[width=\linewidth]{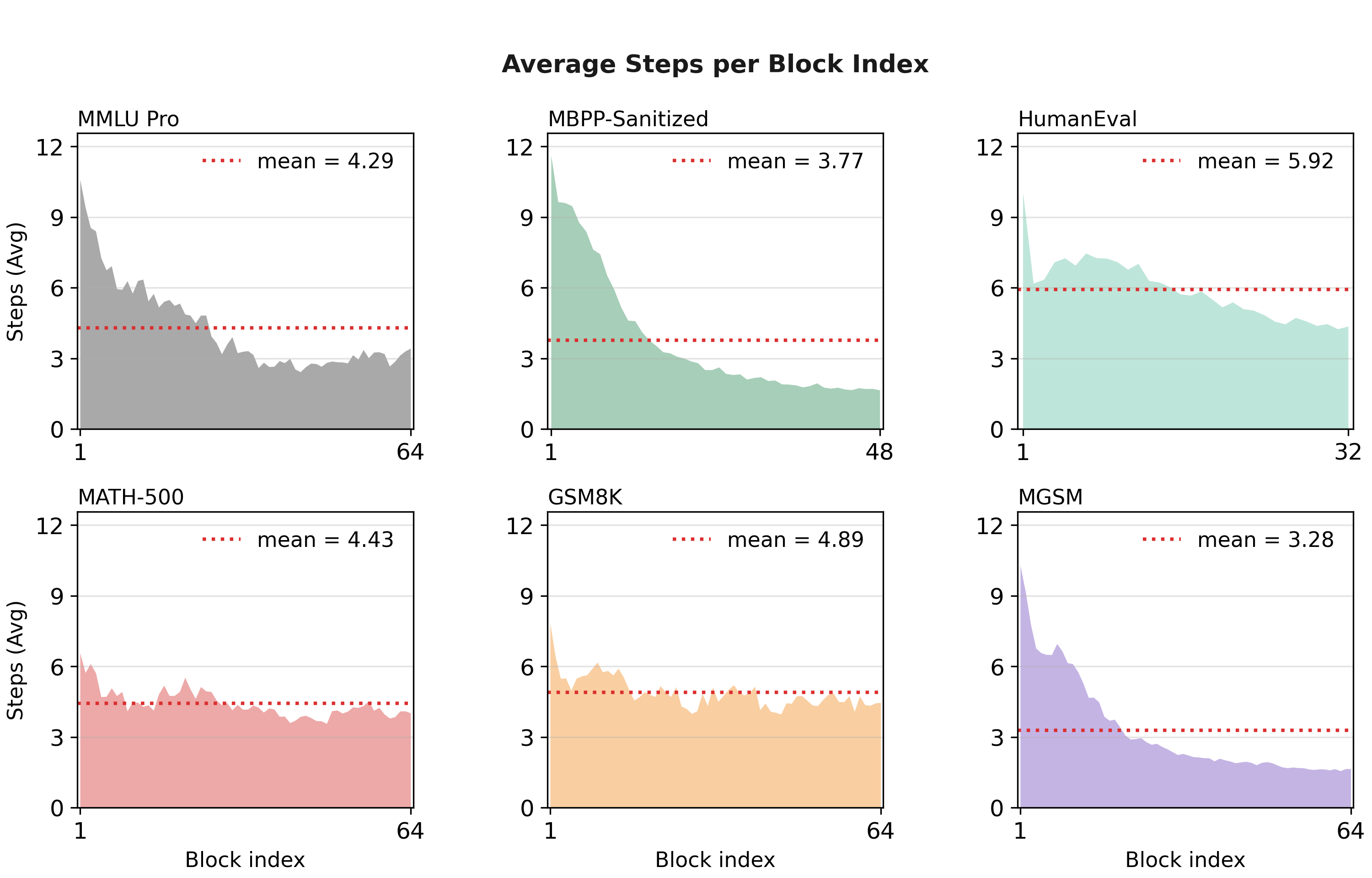}
  \caption{Average diffusion step at which block $b$ completes. Plots terminate at different block indices because different tasks generate different numbers of blocks.}
  \label{fig:appendix-step-x-block}
\end{figure}

\begin{figure}[H]
  \centering
  \includegraphics[width=\linewidth]{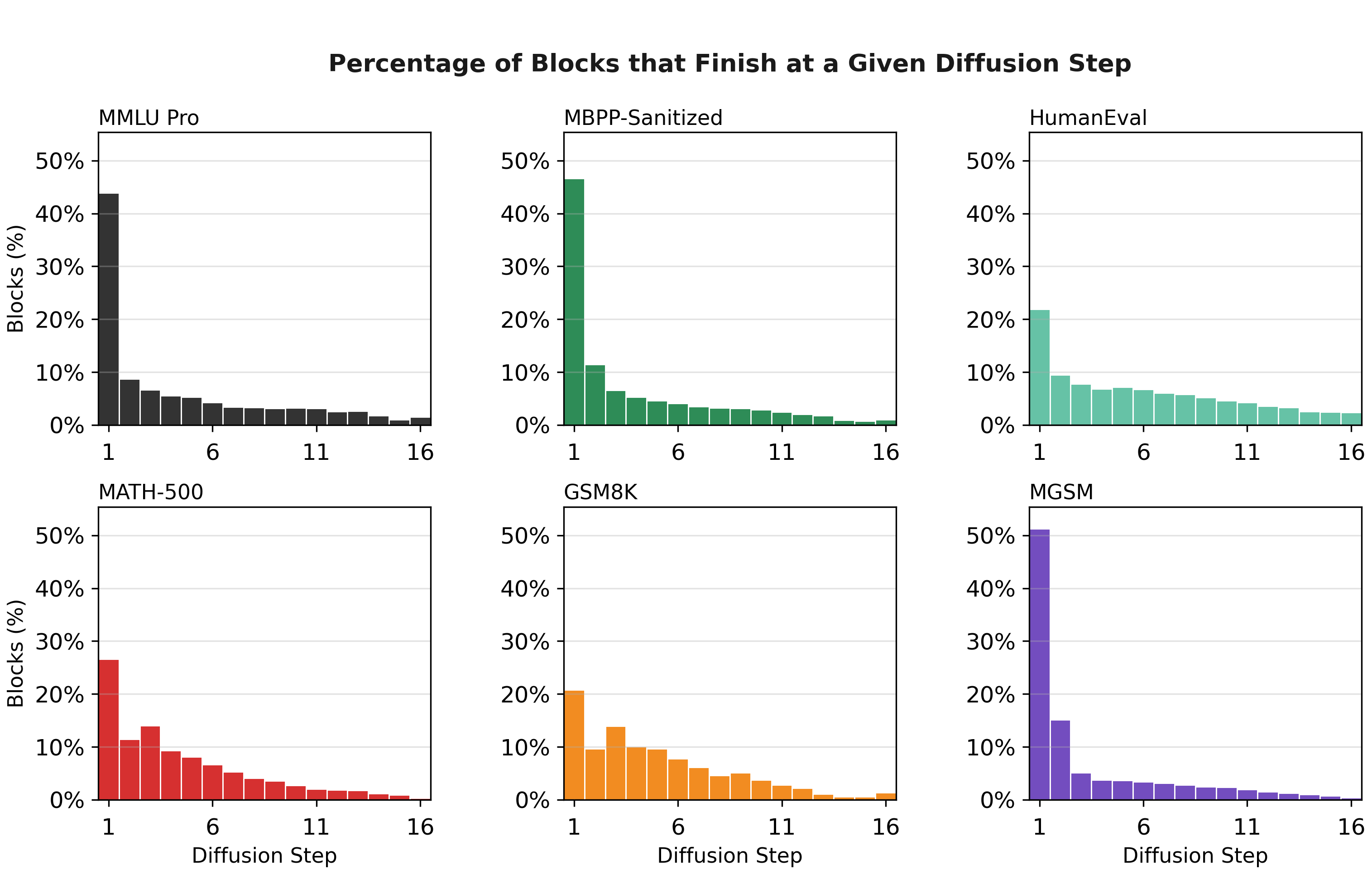}
  \caption{Percentage of generated blocks that complete at each diffusion step across the full task set. Most blocks finish in the first few steps, with task-dependent completion tails.}
  \label{fig:appendix-block-x-step}
\end{figure}

\begin{figure}[H]
  \centering
  \includegraphics[width=\linewidth]{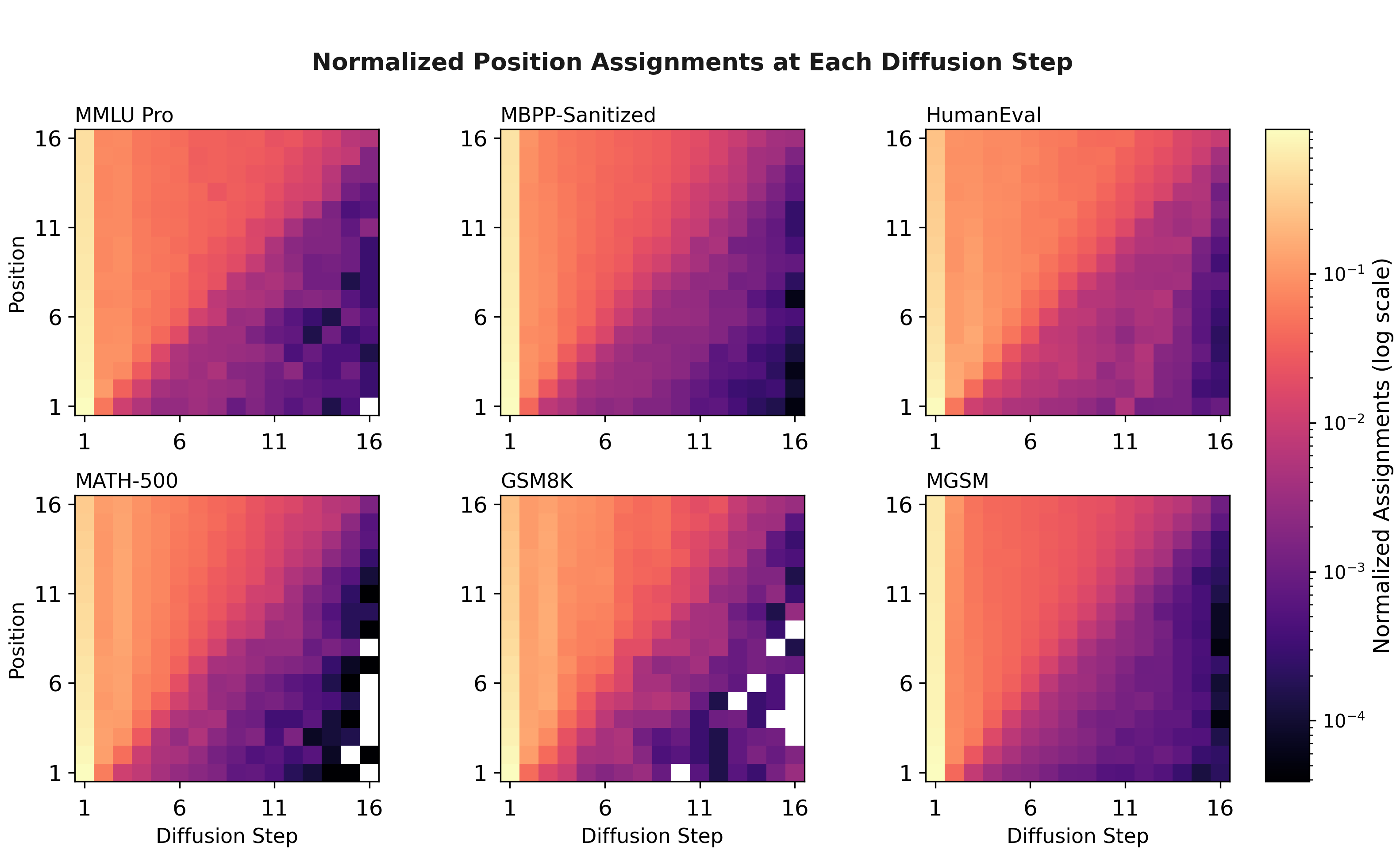}
  \caption{Distribution of token commitments by position and diffusion step across all generated blocks. Earlier positions are typically committed earlier, producing a left-to-right upper-left triangular pattern with task-dependent strength.}
  \label{fig:appendix-position-x-step}
\end{figure}

%% file: references.bib
@article{radford2019language,
  title={Language models are unsupervised multitask learners},
  author={Radford, Alec and Wu, Jeffrey and Child, Rewon and Luan, David and Amodei, Dario and Sutskever, Ilya and others},
  journal={OpenAI blog},
  volume={1},
  number={8},
  pages={9},
  year={2019}
}

@article{grattafiori2024llama,
  title={The llama 3 herd of models},
  author={Grattafiori, Aaron and Dubey, Abhimanyu and Jauhri, Abhinav and Pandey, Abhinav and Kadian, Abhishek and Al-Dahle, Ahmad and Letman, Aiesha and Mathur, Akhil and Schelten, Alan and Vaughan, Alex and others},
  journal={arXiv preprint arXiv:2407.21783},
  year={2024}
}

@article{liu2024deepseek,
  title={Deepseek-v3 technical report},
  author={Liu, Aixin and Feng, Bei and Xue, Bing and Wang, Bingxuan and Wu, Bochao and Lu, Chengda and Zhao, Chenggang and Deng, Chengqi and Zhang, Chenyu and Ruan, Chong and others},
  journal={arXiv preprint arXiv:2412.19437},
  year={2024}
}

@article{austin2021structured,
  title={Structured denoising diffusion models in discrete state-spaces},
  author={Austin, Jacob and Johnson, Daniel D and Ho, Jonathan and Tarlow, Daniel and Van Den Berg, Rianne},
  journal={Advances in neural information processing systems},
  volume={34},
  pages={17981--17993},
  year={2021}
}

@article{sahoo2024simple,
  title={Simple and effective masked diffusion language models},
  author={Sahoo, Subham S and Arriola, Marianne and Schiff, Yair and Gokaslan, Aaron and Marroquin, Edgar and Chiu, Justin T and Rush, Alexander and Kuleshov, Volodymyr},
  journal={Advances in Neural Information Processing Systems},
  volume={37},
  pages={130136--130184},
  year={2024}
}

@article{nie2025large,
  title={Large language diffusion models},
  author={Nie, Shen and Zhu, Fengqi and You, Zebin and Zhang, Xiaolu and Ou, Jingyang and Hu, Jun and Zhou, Jun and Lin, Yankai and Wen, Ji-Rong and Li, Chongxuan},
  journal={arXiv preprint arXiv:2502.09992},
  year={2025}
}

@article{arriola2025block,
  title={Block diffusion: Interpolating between autoregressive and diffusion language models},
  author={Arriola, Marianne and Gokaslan, Aaron and Chiu, Justin T and Yang, Zhihan and Qi, Zhixuan and Han, Jiaqi and Sahoo, Subham Sekhar and Kuleshov, Volodymyr},
  journal={arXiv preprint arXiv:2503.09573},
  year={2025}
}

@article{arriola2025encoder,
  title={Encoder-decoder diffusion language models for efficient training and inference},
  author={Arriola, Marianne and Schiff, Yair and Phung, Hao and Gokaslan, Aaron and Kuleshov, Volodymyr},
  journal={arXiv preprint arXiv:2510.22852},
  year={2025}
}

@article{blakeman2025nvidia,
  title={NVIDIA Nemotron 3: Efficient and Open Intelligence},
  author={Blakeman, Aaron and Grattafiori, Aaron and Basant, Aarti and Gupta, Abhibha and Khattar, Abhinav and Renduchintala, Adi and Vavre, Aditya and Shukla, Akanksha and Bercovich, Akhiad and Ficek, Aleksander and others},
  journal={arXiv preprint arXiv:2512.20856},
  year={2025}
}

@inproceedings{peebles2023scalable,
  title={Scalable diffusion models with transformers},
  author={Peebles, William and Xie, Saining},
  booktitle={Proceedings of the IEEE/CVF international conference on computer vision},
  pages={4195--4205},
  year={2023}
}

@article{ou2024your,
  title={Your absorbing discrete diffusion secretly models the conditional distributions of clean data},
  author={Ou, Jingyang and Nie, Shen and Xue, Kaiwen and Zhu, Fengqi and Sun, Jiacheng and Li, Zhenguo and Li, Chongxuan},
  journal={arXiv preprint arXiv:2406.03736},
  year={2024}
}

@article{chen2023pixart,
  title={{PixArt-$\alpha$: Fast Training of Diffusion Transformer for Photorealistic Text-to-Image Synthesis}},
  author={Chen, Junsong and Yu, Jincheng and Ge, Chongjian and Yao, Lewei and Xie, Enze and Wu, Yue and Wang, Zhongdao and Kwok, James and Luo, Ping and Lu, Huchuan and others},
  journal={arXiv preprint arXiv:2310.00426},
  year={2023}
}

@article{shi2024simplified,
  title={Simplified and generalized masked diffusion for discrete data},
  author={Shi, Jiaxin and Han, Kehang and Wang, Zhe and Doucet, Arnaud and Titsias, Michalis},
  journal={Advances in neural information processing systems},
  volume={37},
  pages={103131--103167},
  year={2024}
}

@misc{loshchilov2019decoupledweightdecayregularization,
      title={Decoupled Weight Decay Regularization}, 
      author={Ilya Loshchilov and Frank Hutter},
      year={2019},
      eprint={1711.05101},
      archivePrefix={arXiv},
      primaryClass={cs.LG},
      url={https://arxiv.org/abs/1711.05101}, 
}

@misc{hu2024minicpmunveilingpotentialsmall,
      title={MiniCPM: Unveiling the Potential of Small Language Models with Scalable Training Strategies}, 
      author={Shengding Hu and Yuge Tu and Xu Han and Chaoqun He and Ganqu Cui and Xiang Long and Zhi Zheng and Yewei Fang and Yuxiang Huang and Weilin Zhao and Xinrong Zhang and Zheng Leng Thai and Kaihuo Zhang and Chongyi Wang and Yuan Yao and Chenyang Zhao and Jie Zhou and Jie Cai and Zhongwu Zhai and Ning Ding and Chao Jia and Guoyang Zeng and Dahai Li and Zhiyuan Liu and Maosong Sun},
      year={2024},
      eprint={2404.06395},
      archivePrefix={arXiv},
      primaryClass={cs.CL},
      url={https://arxiv.org/abs/2404.06395}, 
}

@article{megatron-lm,
  title={Megatron-LM: Training Multi-Billion Parameter Language Models Using Model Parallelism},
  author={Shoeybi, Mohammad and Patwary, Mostofa and Puri, Raul and LeGresley, Patrick and Casper, Jared and Catanzaro, Bryan},
  journal={arXiv preprint arXiv:1909.08053},
  year={2019}
}
